\documentclass{article}

    \usepackage[preprint]{neurips_2021}

\usepackage[utf8]{inputenc} 
\usepackage[T1]{fontenc}    
\usepackage{hyperref}       
\usepackage{url}            
\usepackage{booktabs}       
\usepackage{amsfonts}       
\usepackage{amsthm}
\usepackage{nicefrac}       
\usepackage{microtype}      
\usepackage{xcolor}         
\usepackage{dirtytalk}

\usepackage{physics}
\usepackage{graphicx}

\newtheorem{definition}{Definition}

\title{An automatic differentiation system for the age of differential privacy}

\author{
  Dmitrii Usynin \dag\S, Alexander Ziller \dag, Moritz Knolle, \dag \\ \textbf{Andrew Trask \pounds\ddag, Kritika Prakash\#\ddag} \\ \textbf{ Daniel Rueckert \dag\S, Georgios Kaissis \dag\S\ddag\P} \\ 
\dag Technical University of Munich, \S Imperial College London,\\ \pounds University of Oxford, \ddag OpenMined, \# IIIT Hyderabad \\\P \url{g.kaissis@tum.de}}

\begin{document}

\maketitle

\begin{abstract}
We introduce \textit{Tritium}, an automatic differentiation-based sensitivity analysis framework for differentially private (DP) machine learning (ML). Optimal noise calibration in this setting requires efficient Jacobian matrix computations and tight bounds on the L2-sensitivity. Our framework achieves these objectives by relying on a functional analysis-based method for sensitivity tracking, which we briefly outline. This approach interoperates naturally and seamlessly with static graph-based automatic differentiation, which enables order-of-magnitude improvements in compilation times compared to previous work. Moreover, we demonstrate that optimising the sensitivity of the entire computational graph at once yields substantially tighter estimates of the true sensitivity compared to interval bound propagation techniques. Our work naturally befits recent developments in DP such as individual privacy accounting, aiming to offer improved privacy-utility trade-offs, and represents a step towards the integration of accessible machine learning tooling with advanced privacy accounting systems.   
\end{abstract}

\section{Introduction}
Despite the growing availability of high-performance algorithmic tools for advanced statistical modelling and machine learning, solutions to many of the world’s most important problems require access to sensitive or confidential data. Technologies such as differential privacy can allow drawing insights from such data while objectively allocating and quantifying individual privacy expenditure. Although DP is the gold standard for data protection, its application to everyday ML workflows is \textendash in practice \textendash often constrained. For one, tightly introspecting the privacy attributes of complex models such as deep neural networks can be very challenging. Moreover, substantial expertise is required on the analyst's behalf to correctly apply DP mechanisms to such models. Software libraries \cite{opacus, tfprivacy, holohan2019diffprivlib, googledp} are being developed to alleviate these issues in specific domains such as DP deep learning. They are, however, limited to a small number of programming languages and application programming interfaces (APIs). The democratisation of DP machine learning therefore awaits generic infrastructure, not only compatible with arbitrary workflows, but designed \say{from first principles} to facilitate the implementation of DP. At its core, contemporary ML is based around the manipulation of multidimensional arrays and the composition of differentiable functions, a programming paradigm referred to as \textit{differentiable programming}. Besides deep learning, some of the most successful ML algorithms \cite{chen2016xgboost} and a large number of statistical queries, especially from the domain of \textit{robust statistics}, can be expressed within this paradigm. \textit{Automatic differentiation} (AD) systems are the core of differentiable programming frameworks and are able to track the flow of computation to return precise derivatives with respect to arbitrary computational quantities. Although this functionality may \textendash at first \textendash seem orthogonal to the goals described above, we contend that it is in fact not only highly compatible, but \textit{synonymous} with automatic DP tracking. \par
In the current work, we present \textit{Tritium}, a differentiable programming framework aiming to integrate the requirements of ML and privacy analysis through the use of AD. We recapitulate the link between the \textit{sensitivity} of differentiable queries and the \textit{Lipschitz} constant in section \ref{sec:theory}. We outline our system's implementation in section \ref{sec:impl} and present the substantial improvements in computational efficiency and sensitivity bound tightness in section \ref{sec:experiments}. A discussion of prior work can be found in the appendix.

\section{Theoretical motivation}
\label{sec:theory}
We begin by briefly introducing an interpretation of DP using the language of functional analysis, which forms the theoretical motivation behind our work. We concentrate on the Gaussian mechanism, which forms the basis for private data analysis in high dimensions. Differentially private ML can fundamentally be abstracted as the application of a higher-order function (or \textit{functional}) to private data. This higher-order function (often termed a \textit{DP mechanism}) receives as its input another function (termed a \textit{query}) which has been applied to a private dataset, inspects the query to derive its privacy attributes and modifies it to preserve DP (Definitions \ref{def:DP} and \ref{def:mech}).

\begin{definition}[Query]
A \textit{query} is a function $q: \mathrm{R}^{m \times \mathcal{D}} \rightarrow \mathrm{R}^{n \times \mathcal{D}}$, where $\mathcal{D}$ represents arbitrary (possibly unused) dimensions and $n,m \geq 1$ which receives as input some private dataset $\mathbf{x}, \vert \mathbf{x} \vert \geq 1$ and outputs a result $\mathbf{y}$ representing the result of a computation over $\mathbf{x}$ (e.g. a mean calculation or the output of a neural network). 
\label{def:DP}
\end{definition}

\begin{definition}[DP mechanism]
A \textit{DP mechanism} is a higher-order function $M$ which receives as its input one or more query functions $q_1, q_2, \dots, q_n$ and outputs $M(q_1 \circ q_2 \circ \dots \circ q_n) = q_n(\dots q_2(q_1(\mathbf{x}))) + \xi$, where $\xi \sim \mathcal{N}(0,C)$ and $C$ is selected based on the privacy properties of $q_1 \circ q_2 \circ \dots \circ q_n$.    
\label{def:mech}
\end{definition}
The tight characterisation of these privacy properties is central to enabling privacy expenditure tracking. The effect on inputs on the output of the query functions is reflected in query \textit{sensitivity}. We use the \textit{Lipschitz} constant to reason about sensitivity.

\begin{definition}[Lipschitz constant and sensitivity]
Let $q:X \rightarrow Y$ be a function between metric spaces $X$ and $Y$ with distance metrics $d_X$ and $d_Y$, respectively. Then $q$ is Lipschitz continuous with constant $K_q$ (equivalently, \say{$K$-Lipschitz}) if

\begin{equation}
    d_Y(f(x), f(x')) \leq K ~ d_X(x,x') ~\forall x, x' \in X
    \label{eq:lip}
\end{equation}
The smallest value of $K$ corresponds to the sensitivity of $q, \Delta_2(q)$ \cite{raskhodnikova2016lipschitz}:

\begin{equation}
    \Delta_2(q) = \max_{x,x'} \Vert q(x)-q(x')\Vert_{2}
\end{equation}
where $\Vert \cdot \Vert_2$ is the $L_2$ distance. Recall that in this case, $d_X(x, x')=1$ as $x,x'$ are adjacent, i.e. the Hamming distance between $x$ and $x'$ equals $1$.
Thus for differentiable query functions, $K_q \equiv \Delta_2 (q)$ when $X$ and $Y$ are Euclidean spaces endowed with the $L_2$-norm. Then,
\begin{equation}
K_q  = \sup \Vert \mathcal{J}(q) \Vert_2    
\end{equation}
where $\mathcal{J}$ is the Jacobian matrix (the differential operator).
\end{definition}

This equivalence between Lipschitz constant and query sensitivity allows, in principle, to reason over the privacy attributes of individual query functions and calibrate noise appropriately. Typical functions with globally bounded sensitivity are affine queries or linear functions of the form $q(x) = \alpha x (+ \beta)$, with $K_q = \alpha$. However, queries exist for which the Lipschitz constant is not defined over the entire input domain. One example of such a function is $q(x) = x^2 =x \cdot x$ with $K_q = 2x$, whose sensitivity is \textit{unbounded}, as it depends on the value of $x$. The sensitivity analysis of such queries sometimes therefore depends on (private) properties of the dataset. We term such a case as \textit{data-dependent sensitivity}. Reasoning over sensitivity in such cases is complicated by a requirement to propagate this data dependency effect through function composition. Previous works on Lipschitz analysis of machine learning algorithms \cite{bhowmick2021lipbab, LipOuterBound} achieve this through techniques such as \textit{interval bound propagation} \cite{gowal2019effectiveness}, that carry the bounds on input variables (which, for DP, should be defined in a data-agnostic manner) through the computation flow. This technique can easily be made compatible with \textit{tracing-type} AD systems which are widely used in contemporary machine learning. However, due to well-known limitations of interval arithmetic (such as interval dependency \cite{Krmer2006}) and due to the fact that the Lipschitz constant is defined by inequality, the resulting sensitivity terms may be valid, but too loose to be of any practical utility (e.g. $10^5$). The last challenge relates to the fact that the \textit{actual} effect of an individual's data on the query's output may, in fact, be much smaller than the worst case assumed by the definition, resulting in more noise being added by the mechanism than would be required for the guarantee to hold. Consequently, although the worst-case sensitivity value has typically been used for privacy accounting, newer techniques perform accounting based not only the worst case but combine it with the actual output $L_2$-norm \cite{feldman2020individual}. 

\section{Implementation details}
\label{sec:impl}
Our work presents \textit{Tritium}, an automatic-differentiation-based machine learning and sensitivity analysis system engineered to address the above-mentioned challenges. It consists of the following components:
\begin{enumerate}
    \item A user-facing front-end to specify a query $\mathbf{q} = q_1 \circ \dots \circ q_n$ \textit{abstractly}, i.e. without directly utilising private data during model creation. This is achieved through the utilisation of abstract tensors with pre-defined dimensions. The system creates an optimised computational graph $\mathcal{G}$ based on this specification.
    \item During model specification, the user can impose \textit{bounds} on the quantities (e.g. inputs, weights) used in the model. 
    \item The user selects the desired \textit{privacy parameters}, e.g. $\varepsilon$ and $\delta$ values or a maximum allowed sensitivity.
    \item A \textit{compiler} then emits a program which receives a private dataset and outputs an appropriately privatised result. 
\end{enumerate}
Internally, \textit{Tritium} undertakes the following steps:
\begin{enumerate}
    \item The computational graph $\mathcal{G}$ is compiled into a program which outputs $\mathcal{J}(\mathbf{q})$ with respect to the inputs.
    \item $K_q = \sup \Vert \mathcal{J}(\mathbf{q}) \Vert_2$ is computed given the input bounds.
    \item Finally, $\mathcal{G}$ is compiled into the program described in step (4) above which receives a private dataset $\mathbf{x}$, computes $\mathbf{q}(\mathbf{x})$, potentially \textit{clips} out-of-bound values to preserve the required $K_{\mathbf{q}}$, adds noise $\xi \sim \mathcal{N}(0, C)$ with $C$ proportional to $K_{\mathbf{q}}$ to satisfy the required $\varepsilon$ value for a given $\delta$ and outputs $M(\mathbf{q})$.
\end{enumerate}
This system architecture has several benefits: It avoids utilising private data until the moment the final computation is executed (\textit{data minimisation}). Moreover, it provides a tight sensitivity calculation by optimising the entire query function at once \cite{gouk2021regularisation} instead of the above-mentioned forward-propagation, which can lead to vacuous sensitivity values. Furthermore, it utilises the pre-specified bounds on the input variables to not only enable the calculation of \textit{data-dependent} sensitivity, but also greatly accelerate the process. Moreover, it is \textit{agnostic} to the method used to actually \textit{obtain} the desired sensitivity. For example, \textit{Lipschitz neural network layers} \cite{shavit2019exploring, anil2019sorting} or activation functions with bounded outputs and gradients \cite{papernot2020tempered} can be used for model building, but bounded sensitivity can also be enforced by clipping, as is common in DP-SGD \cite{abadi2016deep}. In addition, the system is able to compute the full Jacobian matrix (which is required in DP-SGD) as well as arbitrary higher-order derivative matrices (which can be used to accelerate the sensitivity computation). Additionally, as the system outputs both the Lipschitz constant and the norm of the outputs, it can be leveraged to provide tighter privacy guarantees through the use of e.g. \textit{individual privacy accounting}, as shown below. Finally, the system is designed to output privatised values by default instead of outputting non-private values and relying on the user to perform an appropriate privatisation step. This can reduce both user workload and the probability of failure due to incorrect application of DP mechanisms on the user's behalf.

\section{Experimental evaluation}
\label{sec:experiments}
\subsection{Exact sensitivity calculations through \textit{a posteriori} optimisation}
To assess the benefits of computing query sensitivity by assessing the entire computational graph at once instead of forward-propagating interval bounds, we constructed a small neural network comprising 4 \textit{linear} layers with \textit{logistic sigmoid} activations and the \textit{binary cross-entropy} cost function. We set the bounds for the neural network weights and the input bounds to the $[0,1]$ interval. We then calculated the sensitivity using two techniques: \textit{Interval Bound Propagation} (IBP \cite{gowal2019effectiveness}) and our proposed method optimising the entire computational graph at once. The estimate of the sensitivity was returned as $[0.0, 22175.37]$ by IBP (which is a valid, but vacuous bound) and as $0.99929$ by \textit{Tritium}. The IBP bounds are also similar to previous work (compare e.g. \cite{zhang2019recurjac}). However, IBP was faster, requiring $103$ ms to return a result, compared with $905$ ms for our technique (excluding compilation time of ca. $3$ s). These results are summarised in Table \ref{tab:table} in the appendix.

\subsection{Compilation improvements}
In this section, we compare the compilation and execution time improvements of \textit{Tritium} to the previously described framework by \cite{ziller2021sensitivity} on neural network architectures with the architecture described above, but with increasing width of linear layers. We recall that the system proposed by the authors of this work relies on a scalar AD implementation and authors report compilation times quadratic in the number of model parameters (that is, around 60 hours for a $2.5$ million parameter model). In contrast, the here-presented implementation utilises the \textit{Aesara} library (a fork of the now-defunct \textit{Theano} framework \cite{theano}) as a computational back-end. This allows both a memory-efficient vectorised execution of tensor operations and a more mature and faster compilation back-end. For all but the smallest architectures, this back-end achieved substantially faster compilation times which were independent of the number of parameters and able to leverage caching to accelerate re-compilation. A similar effect was observed in execution times, where our system achieved considerably higher performance. 
These results are visualised in Figure \ref{fig:perf_comp} in the appendix.

\section{Discussion and conclusion}
We propose \textit{Tritium}, an automatic differentiation-based system for differentially private machine learning. Our framework relies on an interpretation of DP queries and mechanisms through the language of functional analysis, linking them by the definition of Lipschitz continuity. We found the combination of an efficient computational and compilation back-end with the consideration of the entire query function at once to yield both improved performance and tighter sensitivity bounds compared to previous work. Our proposed framework relies on \textit{static graph-based} AD, which can apply specific compiler optimisations to the entire computational graph and is responsible for \textit{Tritium}'s high performance. However, such systems have noteworthy limitations. For instance, the definition of control flow statements is cumbersome and such systems are not well-suited for utilisation with \textit{just-in-time} compilers. Most currently used machine learning frameworks utilise \textit{tracing}/\textit{eager execution} AD back-ends, which, while more user-friendly, cannot always leverage the same optimisations. An alternative AD implementation, \textit{source-to-source translation} can combine the benefits of dynamic graph specification with the high performance and optimisations of static compilation. It forms the basis of a recent paradigm in programming language design (e.g. \cite{saeta2021swift}), attempting to merge a general-purpose programming language with differentiable programming primitives. A limitation of our method stems from the computational hardness of exactly calculating the Lipschitz constant \cite{scaman2018lipschitz}. Our system will output the true bound using \textit{Simplicial Homology Global Optimisation} \cite{endres2018simplicial} if the bound exists and can be found, and the application of constraints can substantially accelerate this process. However, it will output a warning and switch to an approximate algorithm without a guaranteed bound otherwise. If such a bound is undesirable, an alternative technique is the utilisation of model components with known (or manually adjustable) Lipschitz constants, which can allow one to avoid the utility penalty imposed by clipping-based approaches, both enabling the design of algorithms with milder privacy-utility penalties and additionally reaping the benefits of well-defined model sensitivity, such as (certifiable) robustness to perturbations by adversarial samples \cite{lecuyer2019certified}. Moreover, our work serves as a first proof-of-concept for the the design of generic infrastructure exposing familiar APIs to data scientists while automatically tracking privacy loss through the computation flow \cite{trask2021towards}. We view the further development of such systems as an accelerator for the wide-spread adoption of privacy-preserving machine learning algorithms across data-driven research disciplines.  

\bibliographystyle{unsrt}
\bibliography{main}

\begin{thebibliography}{10}

\bibitem{opacus}
{O}pacus {PyTorch} library.
\newblock Available from \href{https://opacus.ai}{opacus.ai}, 2021.

\bibitem{tfprivacy}
{TensorFlow Privacy}.
\newblock Available from
  \href{https://github.com/tensorflow/privacy}{TensorFlow Privacy}, 2021.

\bibitem{holohan2019diffprivlib}
Naoise Holohan, Stefano Braghin, Pól~Mac Aonghusa, and Killian Levacher.
\newblock Diffprivlib: The ibm differential privacy library, 2019.

\bibitem{googledp}
{Differential Privacy}.
\newblock Available from
  \href{https://github.com/google/differential-privacy}{google/differential-privacy},
  2021.

\bibitem{chen2016xgboost}
Tianqi Chen and Carlos Guestrin.
\newblock Xgboost: A scalable tree boosting system.
\newblock In {\em Proceedings of the 22nd acm sigkdd international conference
  on knowledge discovery and data mining}, pages 785--794, 2016.

\bibitem{raskhodnikova2016lipschitz}
Sofya Raskhodnikova and Adam Smith.
\newblock Lipschitz extensions for node-private graph statistics and the
  generalized exponential mechanism.
\newblock In {\em 2016 IEEE 57th Annual Symposium on Foundations of Computer
  Science (FOCS)}, pages 495--504. IEEE, 2016.

\bibitem{bhowmick2021lipbab}
Aritra Bhowmick, Meenakshi D'Souza, and G~Srinivasa Raghavan.
\newblock Lipbab: Computing exact lipschitz constant of relu networks.
\newblock {\em arXiv preprint arXiv:2105.05495}, 2021.

\bibitem{LipOuterBound}
Sungyoon Lee, Jaewook Lee, and Saerom Park.
\newblock Lipschitz-certifiable training with a tight outer bound.
\newblock In H.~Larochelle, M.~Ranzato, R.~Hadsell, M.~F. Balcan, and H.~Lin,
  editors, {\em Advances in Neural Information Processing Systems}, volume~33,
  pages 16891--16902. Curran Associates, Inc., 2020.

\bibitem{gowal2019effectiveness}
Sven Gowal, Krishnamurthy Dvijotham, Robert Stanforth, Rudy Bunel, Chongli Qin,
  Jonathan Uesato, Relja Arandjelovic, Timothy Mann, and Pushmeet Kohli.
\newblock On the effectiveness of interval bound propagation for training
  verifiably robust models, 2019.

\bibitem{Krmer2006}
Walter Kr\"{a}mer.
\newblock Generalized intervals and the dependency problem.
\newblock {\em {PAMM}}, 6(1):683--684, December 2006.

\bibitem{feldman2020individual}
Vitaly Feldman and Tijana Zrnic.
\newblock Individual privacy accounting via a renyi filter.
\newblock {\em arXiv preprint arXiv:2008.11193}, 2020.

\bibitem{gouk2021regularisation}
Henry Gouk, Eibe Frank, Bernhard Pfahringer, and Michael~J Cree.
\newblock Regularisation of neural networks by enforcing lipschitz continuity.
\newblock {\em Machine Learning}, 110(2):393--416, 2021.

\bibitem{shavit2019exploring}
Yonadav Shavit and Boriana Gjura.
\newblock Exploring the use of lipschitz neural networks for automating the
  design of differentially private mechanisms.
\newblock {\em Technical Report}, 2019.

\bibitem{anil2019sorting}
Cem Anil, James Lucas, and Roger Grosse.
\newblock Sorting out lipschitz function approximation.
\newblock In {\em International Conference on Machine Learning}, pages
  291--301. PMLR, 2019.

\bibitem{papernot2020tempered}
Nicolas Papernot, Abhradeep Thakurta, Shuang Song, Steve Chien, and {\'U}lfar
  Erlingsson.
\newblock Tempered sigmoid activations for deep learning with differential
  privacy.
\newblock {\em arXiv preprint arXiv:2007.14191}, 2020.

\bibitem{abadi2016deep}
Martin Abadi, Andy Chu, Ian Goodfellow, H~Brendan McMahan, Ilya Mironov, Kunal
  Talwar, and Li~Zhang.
\newblock Deep learning with differential privacy.
\newblock In {\em Proceedings of the 2016 ACM SIGSAC conference on computer and
  communications security}, pages 308--318, 2016.

\bibitem{zhang2019recurjac}
Huan Zhang, Pengchuan Zhang, and Cho-Jui Hsieh.
\newblock Recurjac: An efficient recursive algorithm for bounding jacobian
  matrix of neural networks and its applications.
\newblock In {\em Proceedings of the AAAI Conference on Artificial
  Intelligence}, volume~33, pages 5757--5764, 2019.

\bibitem{ziller2021sensitivity}
Alexander Ziller, Dmitrii Usynin, Moritz Knolle, Kritika Prakash, Andrew Trask,
  Rickmer Braren, Marcus Makowski, Daniel Rueckert, and Georgios Kaissis.
\newblock Sensitivity analysis in differentially private machine learning using
  hybrid automatic differentiation.
\newblock {\em {ICML Theory and Practice of Differential Privacy Workshop}},
  2021.

\bibitem{theano}
{Theano Development Team}.
\newblock {Theano: A {Python} framework for fast computation of mathematical
  expressions}.
\newblock {\em arXiv e-prints}, abs/1605.02688, May 2016.

\bibitem{saeta2021swift}
Brennan Saeta, Denys Shabalin, Marc Rasi, Brad Larson, Xihui Wu, Parker Schuh,
  Michelle Casbon, Daniel Zheng, Saleem Abdulrasool, Aleksandr Efremov, Dave
  Abrahams, Chris Lattner, and Richard Wei.
\newblock Swift for tensorflow: A portable, flexible platform for deep
  learning, 2021.

\bibitem{scaman2018lipschitz}
Kevin Scaman and Aladin Virmaux.
\newblock Lipschitz regularity of deep neural networks: analysis and efficient
  estimation.
\newblock {\em arXiv preprint arXiv:1805.10965}, 2018.

\bibitem{endres2018simplicial}
Stefan~C Endres, Carl Sandrock, and Walter~W Focke.
\newblock A simplicial homology algorithm for lipschitz optimisation.
\newblock {\em Journal of Global Optimization}, 72(2):181--217, 2018.

\bibitem{lecuyer2019certified}
Mathias Lecuyer, Vaggelis Atlidakis, Roxana Geambasu, Daniel Hsu, and Suman
  Jana.
\newblock Certified robustness to adversarial examples with differential
  privacy.
\newblock In {\em 2019 IEEE Symposium on Security and Privacy (SP)}, pages
  656--672. IEEE, 2019.

\bibitem{trask2021towards}
Andrew Trask and Kritika Prakash.
\newblock Towards general-purpose infrastructure for protecting scientific data
  under study.
\newblock {\em {NEURIPS PPML Workshop}}, 2020.

\bibitem{ziller2021pysyft}
Alexander Ziller, Andrew Trask, Antonio Lopardo, Benjamin Szymkow, Bobby
  Wagner, Emma Bluemke, Jean-Mickael Nounahon, Jonathan Passerat-Palmbach,
  Kritika Prakash, Nick Rose, et~al.
\newblock Pysyft: A library for easy federated learning.
\newblock In {\em Federated Learning Systems}, pages 111--139. Springer, 2021.

\bibitem{hall2021syft}
Adam~James Hall, Madhava Jay, Tudor Cebere, Bogdan Cebere, Koen~Lennart van~der
  Veen, George Muraru, Tongye Xu, Patrick Cason, William Abramson, Ayoub
  Benaissa, et~al.
\newblock Syft 0.5: A platform for universally deployable structured
  transparency.
\newblock {\em arXiv preprint arXiv:2104.12385}, 2021.

\bibitem{yoshida2017spectral}
Yuichi Yoshida and Takeru Miyato.
\newblock Spectral norm regularization for improving the generalizability of
  deep learning.
\newblock {\em arXiv preprint arXiv:1705.10941}, 2017.

\bibitem{gupta2021adaptive}
Varun Gupta, Christopher Jung, Seth Neel, Aaron Roth, Saeed Sharifi-Malvajerdi,
  and Chris Waites.
\newblock Adaptive machine unlearning.
\newblock {\em arXiv preprint arXiv:2106.04378}, 2021.

\bibitem{abuah2021dduo}
Chike Abuah, Alex Silence, David Darais, and Joe Near.
\newblock Dduo: General-purpose dynamic analysis for differential privacy.
\newblock {\em arXiv preprint arXiv:2103.08805}, 2021.

\bibitem{pistone2021identity}
Paolo Pistone.
\newblock From identity to difference: A quantitative interpretation of the
  identity type.
\newblock {\em arXiv preprint arXiv:2107.06150}, 2021.

\end{thebibliography}
\newpage
\appendix

\section{Tables and Figures}
\begin{table}[h]
\centering
\begin{tabular}{@{}ccc@{}}
\toprule
     & Upper bound & Computation time (ms) \\ \midrule
Ours & 0.99929     & 905                   \\
IBP  & 22175.37    & 103                   \\ \bottomrule
\end{tabular}
\caption{Comparison of the sensitivity bounds and computation times for our proposed framework  (\textit{Ours}) vs. Interval Bound Propagation (\textit{IBP}).}
\label{tab:table}
\end{table}

\begin{figure}[h]
\centering
\includegraphics[width=0.5\textwidth]{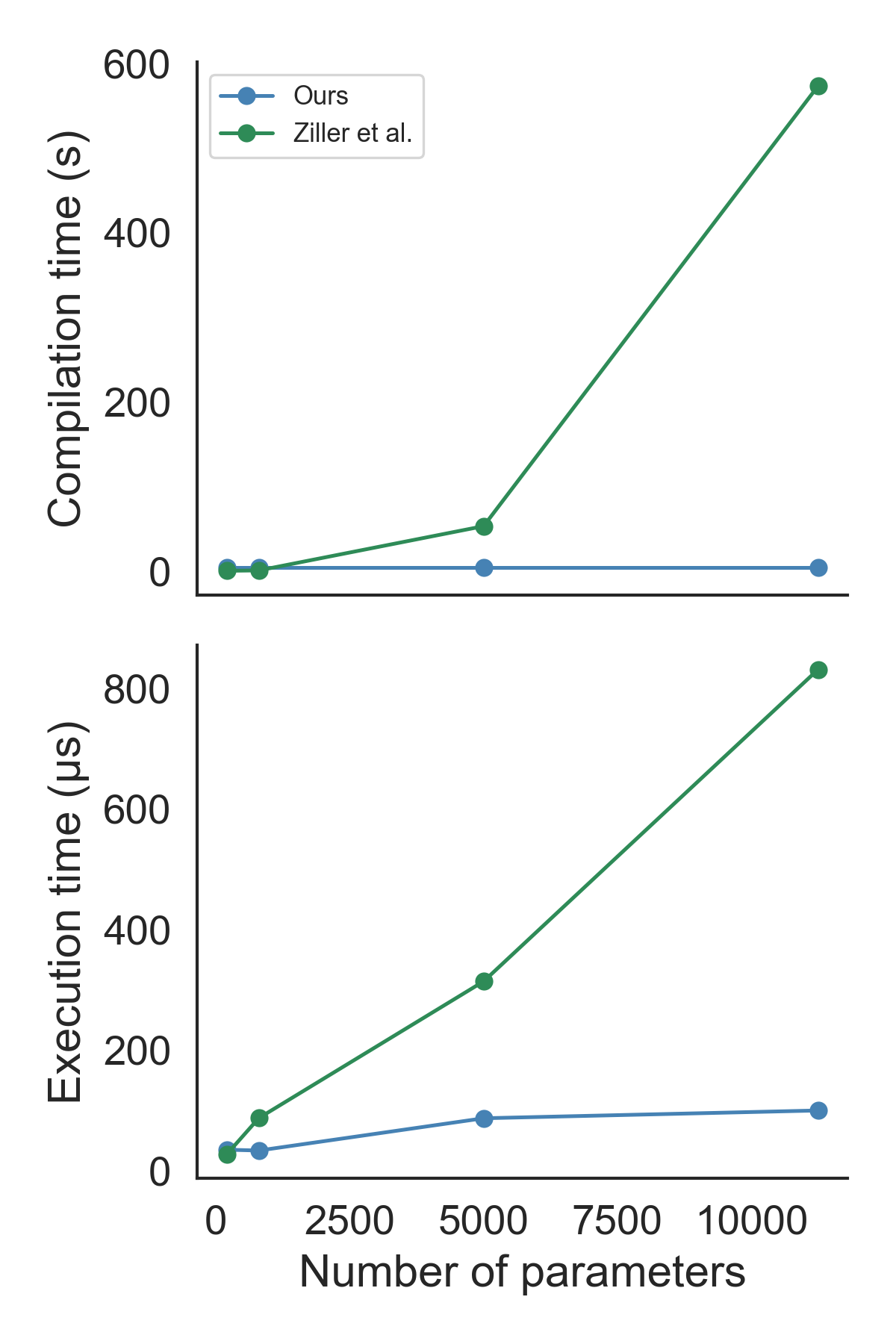}
\caption{Performance comparison between our proposed framework (\textit{Ours}, blue) vs. \cite{ziller2021sensitivity} (green). Compilation times in $s$ and execution times in $\mu s$ are shown for neural network architectures of increasing sizes.}
\label{fig:perf_comp}
\end{figure}

\section{Related works}
Our work can be seen as a natural evolution of the previous study by \cite{ziller2021sensitivity} in the context of AD-based sensitivity analysis for DP machine learning. In comparison to this work, \textit{Tritium} relies on a vectorised, GPU-compatible execution engine and a mature graph compiler which drastically improves performance, as shown in the experimental section. The motivation behind our system's design (described in \cite{trask2021towards}) has recently been implemented in the \textit{PySyft} framework \cite{ziller2021pysyft, hall2021syft} in the form of a so-called \textit{automatic adversarial individual DP accountant}. Here, the sensitivity of statistical database queries is derived and combined with the aforementioned approach by \cite{feldman2020individual} to internally monitor each individual's privacy budget. When the individual privacy budget is exhausted, the individual's data is automatically \textit{filtered}, that is, ejected from the computation in a privacy-preserving manner (imperceptibly to the data scientist executing the query). Our here-presented work is complementary to the PySyft implementation in that it is focused on increased computational speed while not utilising individual DP computations. In the future, we aim to integrate a more advanced AD system into the PySyft codebase. 
The properties of Lipschitz continuous functions have been leveraged in several domains beyond DP. Works such as \cite{shavit2019exploring, yoshida2017spectral, anil2019sorting, zhang2019recurjac} attempt to constrain the Lipschitz constant to reason over and control the properties of neural networks. The utilisation of this approach has been proposed for network certification against adversarial samples \cite{lecuyer2019certified}, whereby a network that is $\epsilon$-certified is provably robust to input perturbations within a norm ball of radius $\epsilon$. Additionally, constraining the Lipschitz constant has been proposed for DP model training, as this allows to calibrate the noise addition based on the bounded Lipschitz constant \cite{shavit2019exploring}. Moreover, certain works \cite{gupta2021adaptive} have addressed the problem of \textit{machine unlearning}, providing methods for a reliable removal of contributions associated with an individual in the context of neural network training. We note that the approach to sensitivity analysis employed in these studies is orthogonal to our work, as our work is compatible with manual sensitivity constraints (such as directly adjusting the Lipschitz constant of neural network layers through appropriate layers as described above, which however may impair their expressivity) but also with sensitivity tracking for privacy loss calculation. Several recent works concentrate on the computation of accurate estimates of the Lipschitz constant mostly focused on \textit{ReLU} networks such as \cite{scaman2018lipschitz, anil2019sorting}, but most of these obtain the upper bounds rather than the exact values of the Lipschitz constant, often resulting in valid, but extremely loose approximations that are not intended to be applied in DP training. A line of work centred on languages for differentially private programming also exists. Among these, the recently proposed \textit{DDuo} framework \cite{abuah2021dduo} performs dynamic sensitivity analysis in the context of DP algorithm specification. As shown above however, this approach does not attempt to derive a tight bound on sensitivity in the setting of unbounded queries, declaring sensitivity as \textit{infinite} and relying exclusively on clipping. More general approaches, including \textit{category-theoretical} views on the intersection of differentiable programming and differential privacy such as \cite{pistone2021identity} have also recently been proposed.

\section{Acknowledgements}
The theoretical underpinnings of this work were conceived in the OpenMined input/output privacy working group meetings. The authors would like to thank the OpenMined community for its support.

\end{document}